\documentclass[10pt,twocolumn,letterpaper]{article}

\usepackage{iccv}
\usepackage{times}
\usepackage{epsfig}
\usepackage{graphicx}
\usepackage{amsmath}
\usepackage{amssymb}

\usepackage{caption} 
\usepackage{booktabs,tabularx}
\newcolumntype{Y}{>{\centering\arraybackslash}X}
\usepackage{subcaption}
\usepackage{multirow}

\usepackage[breaklinks=true,bookmarks=false]{hyperref}

\iccvfinalcopy 

\def\iccvPaperID{****} 

\ificcvfinal\fi

\begin{document}
\title{Slender Object Detection: Diagnoses and Improvements}

\newcommand{\A}{\mathbb{A}}
\newcommand{\B}{\mathbb{B}}
\newcommand{\C}{\mathbb{C}}
\newcommand{\D}{\mathbb{D}}
\newcommand{\E}{\mathbb{E}}
\newcommand{\F}{\mathbb{F}}
\newcommand{\G}{\mathbb{G}}
\newcommand{\HH}{\mathbb{H}}
\newcommand{\I}{\mathbb{I}}
\newcommand{\J}{\mathbb{J}}
\newcommand{\K}{\mathbb{K}}
\newcommand{\LL}{\mathbb{L}}

\newcommand{\FE}{\mathcal{FE}}
\newcommand{\LC}{\mathcal{IP}}
\newcommand{\FA}{\mathcal{FA}}
\newcommand{\PP}{\mathcal{DF}}
\newcommand{\LF}{\mathcal{LF}}
\newcommand{\LA}{\mathcal{LA}}
\newcommand{\GR}[1]{(\color{green}{+#1}\color{black})}
\newcommand{\New}{\color{black}}
\newcommand{\NewEnd}{\color{black}}

\author{Zhaoyi Wan\textsuperscript{\rm 1}\footnotemark[\value{footnote}]\thanks{Authors contribute equally},
Yimin Chen\textsuperscript{\rm 2}\footnotemark[\value{footnote}],
Sutao Deng\textsuperscript{\rm 2}\footnotemark[\value{footnote}],
Kunpeng Chen\textsuperscript{\rm 3},
Cong Yao\textsuperscript{\rm 3},
Jiebo Luo\textsuperscript{\rm 1},\\
\textsuperscript{\rm 1}University of Rochester,
\textsuperscript{\rm 2}Beihang University,
\textsuperscript{\rm 3}Megvii\\
i@wanzy.me, jluo@cs.rochester.edu}
\maketitle
\vspace{-0.2in}

\begin{abstract}
\vspace{-0.1in}
In this paper, we are concerned with the detection of a particular type of objects with extreme aspect ratios, namely \textbf{slender objects}.
In real-world scenarios, slender objects are actually very common and crucial to the objective of a detection system.
However, this type of objects has been largely overlooked by previous object detection algorithms.
Upon our investigation, for a classical object detection method, a drastic drop of $18.9\%$ mAP on COCO is observed, if solely evaluated on slender objects.
Therefore, we systematically study the problem of slender object detection in this work.
Accordingly, an analytical framework with carefully designed benchmark and evaluation protocols is established, in which different algorithms and modules can be inspected and compared.
\New
Our study reveals that effective slender object detection can be achieved ~\textbf{with none of} (1) anchor-based localization; (2) specially designed box representations. Instead, \textbf{the critical aspect of improving slender object detection is feature adaptation}.
It identifies and extends the insights of existing methods that are previously underexploited.
Furthermore, we propose a feature adaption strategy that achieves clear and consistent improvements over current representative object detection methods.
\NewEnd
\end{abstract}

\vspace{-0.2in}
\section{Introduction}

As a fundamental task in computer vision that draws considerable research attention from the community, object detection~\cite{rcnn, fasterrcnn, YOLO} has made substantial progress in recent years.
As the needs of real-world applications in a wide variety of scenarios arise~\cite{widerface, aerial}, the significance of research regarding a particular topic elevates. The works on improving specific aspects~\cite{relation, scale} of object detection, such as detecting dense objects~\cite{florence2018dense, densebox} and small objects~\cite{augmentation}, boost the practical value of object detection and consequently inspire further advances~\cite{cascade, augfpn}.

\begin{table}[tp]
\centering
\vspace{-0.1in}
\caption{$mAP(\%)$ gap between tall and wide objects on COCO. Objects are grouped by the width/height ratio $r_b$ of bounding boxes, where XT=extra tall; T=tall; M=medium; W=wide; XW=extra wide, as defined by \cite{diagnose}.}
\vspace{-0.1in}
\begin{tabularx}{1.0\linewidth}{@{}l*{6}X@{}}
\toprule
Method        & all   &  XT  & T & M & W & XW  \\ \midrule
RetinaNet     &  36.4  & 19.2 & 26.8 &   38.1    &  24.6   &  12.7     \\
Faster        &  37.9  & 23.3 & 31.4 &   39.0    &  26.1   &  16.8     \\
   \bottomrule
\end{tabularx}
\vspace{-0.2in}
\label{tab:aspect-ratio-gap}
\end{table}

While a large portion of the problems have been well investigated and numerous new ideas have been proposed, grand challenges remained in object detection. \cite{retina} propose the focal loss to tackle dense object detection and prompt it to become a common practice for classification loss in object detection.
Object scale has been widely considered in model design, as various detection paradigms~\cite{rcnn}, augmentation schemes~\cite{augmentation}, and modules~\cite{sod} are proposed to improve small object detection. Such insightful works propel object detection methods to transfer from academic research to a wide variety of real-world applications~\cite{plastics, db}. Despite such progress in object detection, one significant problem has not been formally explored in previous works. 

This work is dedicated to studying the problem of slender object detection. From the perspective of academic research, the distinctive properties of slender objects pose special challenges, which give rise to research topics of scientific value. From the perspective of application, once slender objects can be well-handled, the practical utility of object detection systems will become higher.

\begin{figure*}[tp]
\vspace{-0.2in}
\centering
\includegraphics[width=0.8\textwidth]{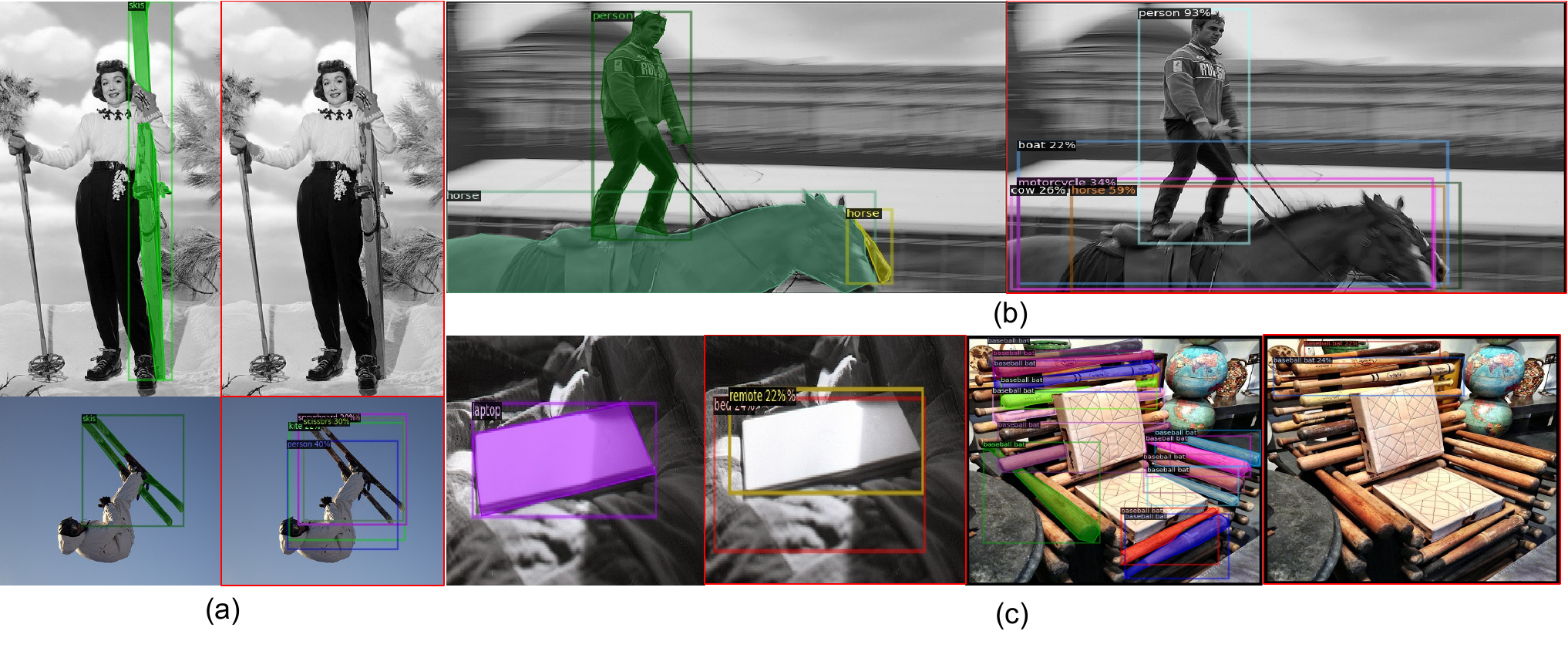}
\vspace{-0.2in}
\caption{Slender object examples. Images marked with red borders visualize detection results.
The ski in the top left is missed by the detector due to anchor mismatching.}
\vspace{-0.2in}
\label{fig:examples}
\end{figure*}

Inspired by previous works~\cite{diagnose,rethinking,survey}, which provide in-depth ablations, analyses, and insights regarding object detection algorithms, we start with diagnosing and analyzing existing methods for object detection.
Specifically, we address issues in slenderness definition, feasible evaluation protocols, and data bias neutralization. A unified analytical framework is established, plus a standard detection pipeline, to dissect and compare different approaches in a clear and fair manner.
We make it convenient to identify the key factors of previous methods, effective choices for model design, and potential directions for improvement.
Key findings related to effective slender object detection, which is proven practicable without anchor-based localization nor specialized box representations, are presented in Sec.~\ref{dissections}.

Beyond diagnoses and analyses, we further propose strategies to boost the detection of slender objects. In particular, a generalized feature adaption module, called self-adaption, is introduced. In addition, we show potential trade-off measures between the detection of slender and regular objects. According to the quantitative experiments (see Sec.~\ref{improvements}), the proposed feature adaptation has proven effective for slender objects while also working well for regular objects (see Tab.~\ref{tab:backbones}).

In summary, the main contributions of this paper are as follows:
\New
\begin{itemize}
    \item We are the first to formally investigate the problem of slender object detection, which is important but largely overlooked by previous works. Issues preventing from in-depth study in slender object detection are addressed, including definition, metrics, and evaluation bias against slender objects. 
    \item We construct an analytical framework for rigorously diagnosing different object detection methods. With this framework, a series of key insights and valuable findings, which may inspire other researchers in the field of object detection, is derived.
    \item We identify the feature adaption module as a key factor for the improvement of slender object detection. A generalized feature adaption module, called self-adaption, is devised. 
\end{itemize}
\NewEnd

\section{Preliminary Assessment}\label{sec:assessment}

\begin{figure}[!htbp]
    \vspace{-0.1in}
    \begin{subfigure}[h]{\columnwidth}
        \centering
        \includegraphics[width=0.75\columnwidth]{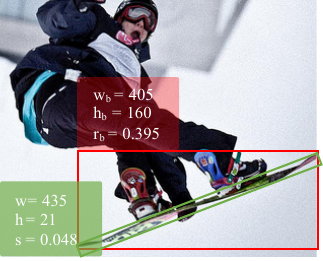} 
    \label{fig:def}
    \end{subfigure}
    \begin{subfigure}[t]{\columnwidth}
        \centering
        \includegraphics[width=0.7\columnwidth]{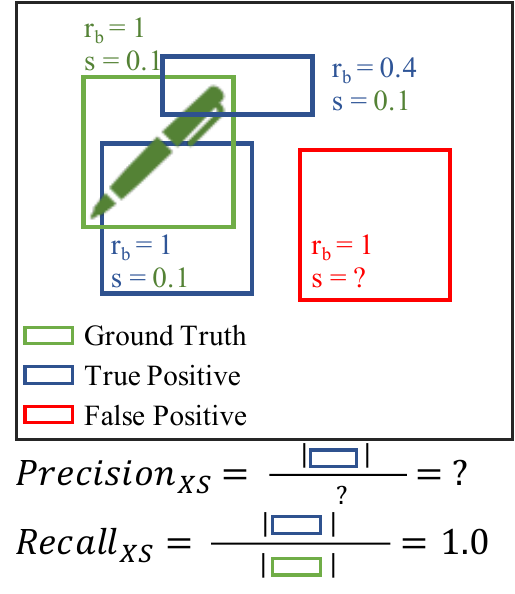}
    \end{subfigure}
    \vspace{-0.3in}
    \caption{\textit{Top:} Aspect ratio from bounding box (red) and slenderness from oriented box (green). \New\textit{Bottom:} Precision for a particular category (XS in the figure) is undefined since the number of false positives can not be obtained. Each GT can match only one positive detection in evaluation.\NewEnd} \label{fig:def}
    \vspace{-0.2in}
\end{figure}

In this section, we will provide an overview of slender object detection and conduct a preliminary assessment on existing methods.
As shown in Fig.~\ref{fig:examples}, slender objects in images can be roughly categorized into three groups: 
\textbf{Distinct slender objects} are those  that are intrinsically slender in shape, such as ski, forks, and bats.
Regular objects may also appear slender in images because of \textbf{occlusion and truncation} (top right in Fig.~\ref{fig:examples}).
In addition, some \textbf{thin plates} in the real world may appear slender from certain viewing angles, e.g., books and tables.
Different categories of objects exhibit different characteristics but may also share some properties in common.
We analyze typical errors by previous methods for these different categories, and accordingly draw unified conclusions regarding slender objects.

\subsection{Definition of Slenderness}\label{definition}

For targeted evaluation and analyses, we need to estimate the slenderness of objects.
In the context of object detection where bounding boxes are the common representation of objects, slenderness can be approximately computed from the width $w_b$ and height $h_b$ of axis-aligned bounding boxes as $r_{b} = w_{b} / h_{b}$.
This formula is specifiable for both the ground truth and detection results, thus being fully applicable to existing evaluation protocols, e.g., mean average precision (mAP) and mean average recall (mAR).
However, the deviation of $r_b$ is obviously inaccurate for oriented slender objects as illustrated in Fig.~\ref{fig:def}. 
It would mistake oriented slender objects as regular objects and in consequence underestimate the gap between regular and slender objects.
The more accurate approach is to find a rotated box which covers the object with the minimal area (green box in Fig.~\ref{fig:def} top), and compute the slenderness $s$ as:
\begin{equation}
    s = \min(w, h) / \max(w, h).
\end{equation}
$w$ and $h$ are the width and height of the minimum-area rectangle.
For the convenience of comparison, we refer to objects with $s < 1/5$, $1/5 < s < 1/3$, $s > 1/3$ as extra slender (XS), slender (S), and regular (R), respectively.

\New
\subsection{Evaluation Metrics}

In addition to the overall mAP that is the common metric for object detection, evaluation over particular aspects, e.g., areas and object class, is desired to diagnosis errors. An example is the evaluation over scales of COCO~\cite{coco} competition.
Meanwhile, the situation is different in respect of object slenderness. Although we can always evaluate the overall mAP, particular mAP over slenderness is undefined.

Given ground truth objects $G$, detections $D$ with confidence scores $S$ produced by an algorithm, mAP is the mean over categories w.r.t average precision over all recalls. Formally, the precision and recall of a category $k$ is defined as a fraction of true positives over predicted positives:
\begin{equation}
    \begin{split}
        \mathcal{P}\left(t, k\right) &= \frac{||D(t,k) \cap G(k)||}{||D(t, k)||},\\
        \mathcal{R}\left(t, k\right) &= \frac{||D(t,k) \cap G(k)||}{||G(t, k)||},
    \end{split}
\end{equation}
where $t$ is the confidence threshold that filters out detections with lower confidence scores. As shown in Fig.~\ref{fig:def} bottom, the current practice of object detection is to represent objects as bounding boxes that leave slenderness unknown. Thus, precision for a particular category is undefined.
\begin{figure}[!t]
\vspace{-0.1in}
\centering
\begin{subfigure}[b]{\columnwidth}
    \centering
    \includegraphics[width=\textwidth]{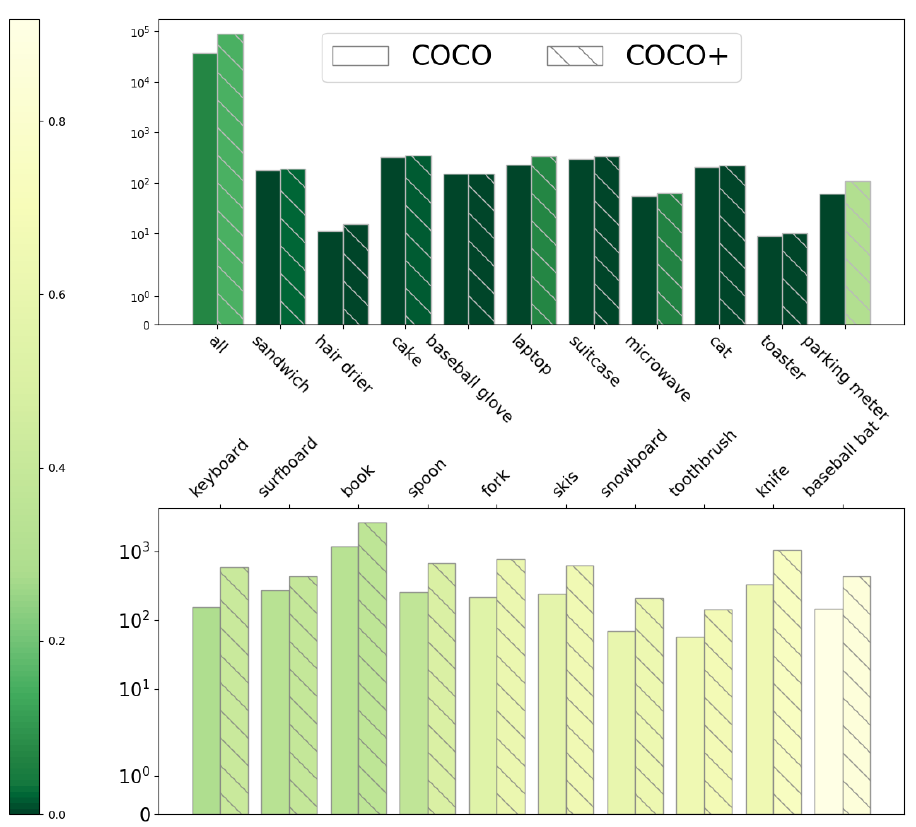}
\end{subfigure}

\begin{subtable}[h]{\columnwidth}
\centering
\begin{tabularx}{1.0\linewidth}{@{}l*{5}X@{}}
\toprule
\multirow{2}{*}{dataset}        &  \multirow{2}{*}{\#images}   &  \multicolumn{3}{c}{\#instances}  \\ \cline{3-5}
                     &             & XS            &    S & R \\ \midrule
COCO        &  5000  & 517 (2.4\%) & 2657 (12.4\%) & 18274 (85.2\%)   \\ \midrule
COCO$^+$       &  23058 & 3486 (5.7\%) & 22343 (36.8\%) & 34964 (57.5\%)  \\
   \bottomrule
\end{tabularx}
\end{subtable}
\vspace{-0.1in}
\caption{Number of instances of different object categories in COCO and COCO$^+$ validation set. Clearly COCO$^+$ is more neutralized in terms of slenderness.}
\vspace{-0.2in}
\label{fig:numbers}
\end{figure}

In contrast, the slenderness category of ground truth can be assigned to matched true positives, making recall feasible for evaluating slender object detection:
\begin{equation}
    \begin{split}
        AR_k &= \int_0^1\mathcal{R}(t, k)dt,\\
        mAR &= \frac{1}{K}\sum_kAR_k.
    \end{split}
\end{equation}
In this paper, we use mAR shown above as the metric for particularly benchamrking slender object detection. Analogically to mAP, the amount of detections for each image is limited to $100$ for evaluation. We also report overall mAP on datasets to compare with existing literature.
\NewEnd

\subsection{Data Bias Neutralization}\label{data}

As mentioned above, we rely on precise boundaries of objects to estimate their slenderness, which is not feasible with conventional axis-aligned bounding box annotations in object detection.
The COCO dataset~\cite{coco}, one of the most popular datasets in recent research of object detection, provides pixel-level segmentation labels.
It is a large-scale dataset collected for object detection and related tasks, e.g., keypoint detection and panoptic segmentation.

However, COCO is biased regarding slender objects and not sufficient for evaluating slender object detection by itself.
The data distribution of COCO is visualized in Fig.~\ref{fig:numbers}, where more than 85\% of objects are regular. The dominant proportion in the dataset implicitly forces the current evaluation to favor regular objects over slender objects.
As shown in Fig.~\ref{fig:slender-performances}, the overall mAR in COCO is close to that of regular objects.
Such a bias against slender objects can be mitigated by extending the validation set of COCO.

We incorporate slender objects from another dataset, Objects365~\cite{obj365}, to complement COCO.
Objects365 is a dataset aiming at object detection in the wild, containing 38k validation images sharing similar characteristics with COCO.
In contrast to COCO which provides detailed boundaries of objects, Objects365 annotates objects with axis-aligned bounding boxes.
We use a top-performing instance segmentation model by \cite{cascade} with a ResNeXt152~\cite{resnext} backbone to generate polygon borders of objects.
Given ground truth bounding boxes during inference, the produced masks are accurate for slenderness estimation.
The procedure and examples of polygon generation are shown in Appendix B in the supplementary material.
According to the slenderness estimated from generated borders, we select images containing extra slender  objects in Objects365 to mix with the COCO validation set, creating COCO$^+$.
As shown in Fig.~\ref{fig:numbers}, the number of slender objects in COCO$^+$ is 8 times more than COCO, thus mitigating the bias against slender objects.
Experimental validation shown in Fig.~\ref{fig:slender-performances} verifies that COCO$^+$ is fairly balanced since the overall mAR is closer to the average of mAR on extra slender objects and mAR of regular objects.

\subsection{Error Analysis}\label{intuitive-analysis}

\begin{figure}[!tbp]
\vspace{-0.2in}
\centering
\includegraphics[width=0.9\columnwidth]{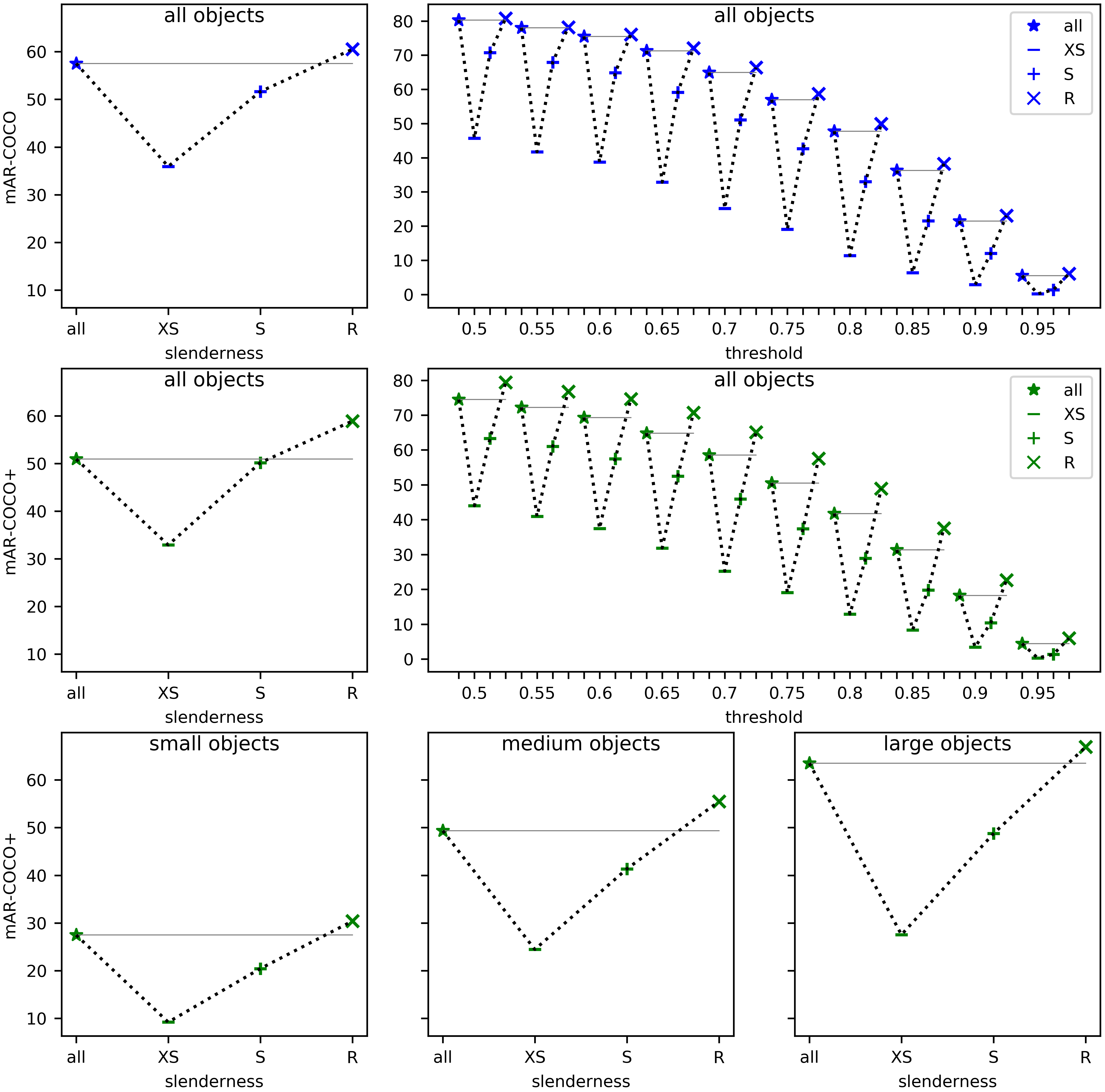}
\vspace{-0.1in}
\caption{
Despite across-the-board performance drop on slender objects, COCO$^+$ is fairly balanced.}
\vspace{-0.2in}
\label{fig:slender-performances}
\end{figure}

Using the evaluation protocols and data, we 
assess the problem by observing the evaluation on a representative method~\cite{retina}.
Models we implemented in this paper are built upon ResNet~\cite{resnet} backbones with FPN~\cite{fpn} and trained on the COCO training set with a 1x schedule.
To make the baseline for experiments, 
we also provide evaluation results on COCO validation set.

\New
The evaluation results are shown in Fig.~\ref{fig:slender-performances}.
It is noteworthy that detection mAR is inversely proportional to object slenderness, with a gap of 19.3\% between XS and R objects.
This correlation is consistent with different data sets, IoU thresholds, and object areas. It verifies that the detection of slender objects is more challenging. In consideration of the notable overlap between slender and small objects, we separately evaluate objects with different areas.
As shown in the last row of Fig.~\ref{fig:slender-performances}, mAR on slender objects are consistently worse than regular objects with a large gap, regardless of the area of objects.
The gap is surprisingly more notable for large objects, due to the increase of challenges in estimating object sizes.

Counterintuitively, accurately classifying slender objects is more difficult than locating them. As shown in Fig~\ref{fig:slender-performances}, evaluation on lower IoU threshold ($<0.7$), which tolerate inaccurate localization results more, bears more significant performance difference. We assume the cause is that bounding boxes of slender objects usually contains more background (see Fig.~\ref{fig:def} for an example), and validate it in latter experiments.
\NewEnd


\begin{table}[tbp]
\centering
\vspace{-0.2in}
\caption{RetinaNet with different sampling rates during training.}\label{tab:data-rate}
\vspace{-0.1in}
\begin{tabularx}{1.0\linewidth}{@{}l*{7}X@{}}
\toprule
\multicolumn{3}{c}{sample rate}  & \multicolumn{4}{c}{COCO$^+$ mAR}  & COCO\\ \cline{1-7}
XS & S & R &   all & XS   &    S & R    &  mAP \\ \midrule
1 & 1 & 1  &  48.4 & 23.4 & 37.8 & 53.9 & 36.4  \\
3 & 2 & 1  &  48.4 & 25.4 & 38.2 & 53.0 & 36.0  \\
10 & 5 & 1 &  48.0 & 25.8 & 37.9 & 52.8 & 35.0  \\
   \bottomrule
\end{tabularx}
\vspace{-0.2in}
\end{table}

An intuitive alleviation of the problems on slender object detection is to increase the sample rate of slender objects in the dataset during training. Its validation is shown in Tab~\ref{tab:data-rate} with a classical baseline RetinaNet~\cite{retina}.
It demonstrates the change of sampling rate in the training data as a trade-off between the effectiveness on regular and slender objects.
Accompanying the increase of slender and extra slender mAR, regular mAR drops.
What we concern more is that, when the sample rates of slender objects continue to increase, the drop of overall performance is also escalated.
Therefore, besides data sampling, we conduct further investigation on models.


\section{Problem Dissection} \label{dissections}

\New
So far, we have addressed issues regarding slender object detection that prevent rigorous investigation of the problem.
An evaluation with the conducted protocol is performed on several representative detection methods and shown in Tab.~\ref{tab:backbones}.
As we can observe from the table, the model advantages over object slenderness fluctuates, making the problem beyond the scope of direct comparison of existing methods. For example, FCOS~\cite{fcos} achieves better detection performance for regular (R) objects, but is limited in detecting slender objects in comparison with RetinaNet. The similar pattern exists between RetinaNet and FasterRCNN.

This problem is concealed by the evaluation bias against slender objects described in Sec.~\ref{intuitive-analysis} and revealed by the proposed particular evaluation.
However, new approaches in object detection are usually brought in with multiple alternations. Their ablation experiments are conducted in divert environments, e.g., backbones~\cite{cornernet, centernet}, training protocols~\cite{cascade, DETR}, thus discouraging direct comparison from recognizing true insights for effective slender object detection. In this section, we first unify main-stream detectors into standard stages, and then combine the proposed evaluation protocols to conduct experiments with controlled variables to reveal the impact of different modules.
\NewEnd




\subsection{Standard Object Detection Stages}\label{descriptive-framework}

\begin{figure}[tb]
\vspace{-0.1in}
\centering
\includegraphics[width=\columnwidth]{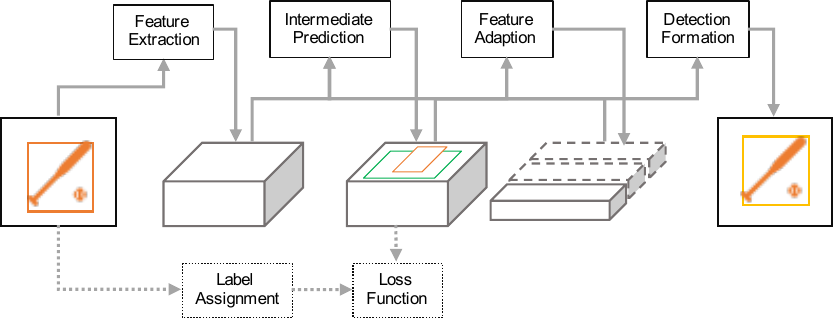}
\vspace{-0.2in}
\caption{Illustration of the decomposed stages of object detection. Note that a given stage can be performed more than once to form the actual pipeline in certain detectors.}
\vspace{-0.2in}
\label{fig:stages}
\end{figure}

Basically, the task of object detection is composed of two subtasks, localization and classification.
A typical detection method localizes and classifies object regions from rich feature maps extracted by neural networks in pyramid resolutions.
Some of existing methods~\cite{YOLO, SSD} directly perform localization and classification on extracted features, and another groups of methods apply feature adaption, e.g. ROI Pooling~\cite{fasterrcnn}, according to coarsely localized object proposals.
They are also referred to as one-stage and two-stage methods in some literature, respectively.
For dense detection, post-processing such as redundancy removal is required for most detectors, after which a set of final object detection is formed.

Deriving from the existing methods, four standard stages of object detection can be defined as follows.
\begin{enumerate}
    \item \textbf{Feature Extraction ($\FE$)} extracts features from the input image to form a high dimensional representation. As deep CNNs~\cite{resnet,resnext} and their variants~\cite{fpn} significantly improve the capability of detectors~\cite{cornernet, centernet}, experimental comparison is usually conducted on the same backbones.
    \item \textbf{Intermediate Prediction ($\LC$)} localizes and/or classifies object regions. They can be assembled together or performed separately. Researchers have developed different paradigms for localization in recent works, where it can be modeled as a regression problem~\cite{retina} or a verification problem~\cite{cornernet}. 
    \item \textbf{Feature Adaptation ($\FA$)} adapts feature maps using intermediate localization results or directly from features for refined prediction. It usually exploits coarse estimation of object regions, i.e. proposals~\cite{rcnn}, as regions of interest to concentrate on objects for refining classification and localization. Note both $\FA$ and $\LC$ can be utilized multiple times~\cite{cascade}.
    \item \textbf{Detection Formation ($\PP$)} forms the final results by removing redundant predictions, filtering low-confidence objects, etc. A common detection formation for object detection is non-maximum suppression (NMS) and its successors.
    Recently, \cite{DETR} propose end-to-end prediction of object localization and classes, resulting in simplified detection formation.
\end{enumerate}

\begin{table}[tp]
\vspace{-0.2in}
\centering
\caption{Evaluation of representative models on slender object detection metrics.}
\vspace{-0.1in}
\begin{tabularx}{1.0\linewidth}{
    p{1.6cm} @{}l*{6}X@{}}
\toprule
\multirow{2}{*}{baseline} & \multirow{2}{*}{backbone} & \multicolumn{4}{c}{COCO$^+$ mAR} & COCO   \\ \cline{3-6}
& & all & XS & S & R & mAP \\ \midrule
RetinaNet &  ResNet-50      & 48.4 & 23.4 & 37.8    &  53.9   &  36.4     \\
RetinaNet &  ResNet-101     & 50.5 & 23.6 & 39.7    &  56.0   &  38.7     \\ \midrule
Faster    &  ResNet-50      & 47.8 & 25.9 & 37.7    &  51.5   &  37.8     \\
Faster    &  ResNet-101     & 48.2 & 25.9 & 37.8    &  52.1   &  38.1     \\ \midrule
FCOS      &  ResNet-50      & 48.7 & 23.2 & 37.9    &  54.4   &  37.6     \\
FCOS      &  ResNet-101     & 50.1 & 24.9 & 39.7    &  55.7   &  40.1     \\
   \bottomrule
\end{tabularx}
\vspace{-0.2in}
\label{tab:backbones}
\end{table}

In addition to these stages that are required for both training and inference, label assignment and loss function identify the criterion for the training procedure.
\textbf{Loss Function ($\LF$)} acts as the optimization target during the training of detectors. It consists of the loss function for classification, where focal loss is dominant, and the loss function for localization, where smooth l1 loss and gIoU loss are preferred choices.
\textbf{Label Assignment ($\LA$)} fills the gap between the optimization target and network outputs. It assigns labels to prediction results, making the model directly trainable.
Label assignment is still under active investigation as it is related to localization and classification representation of objects.
By standardizing stages that identifies a detector, we eliminate undesired variance and conduct rigorous experiments in next sections and believe it will provide a guidance of fair ablation study for the community.

\subsection{Component Inspection}\label{component-inspectino}
\New
Under the analytical framework, we recall typical errors of slender object detection revealed in Sec.~\ref{intuitive-analysis} to inspire our inspection.
One of the major errors of slender object detection is related to \textbf{distinct slender objects} introduced in Sec.~\ref{sec:assessment} and shown in Fig.~\ref{fig:examples}a.
Vertical and horizontal slender objects can be improperly assigned by the IoU matching between the bounding box and pre-defined anchors during training.
In consideration of literate arguing the role of anchors~\cite{guided-anchor, zhong2020anchor}, we introduce the first conjecture to verify: Anchor-based localization is critical for slender object detection.
On the other hand, we also demonstrate the goal of object detection, representing objects in bounding boxes, contradicts with the nature of slender objects. We conjecture works in novel box representations~\cite{reppoints, cornernet} may alleviate this problem and verify it using our analytical framework.


\begin{figure}[tbp]
    \vspace{-0.2in}
    \centering
    \includegraphics[width=0.9\columnwidth]{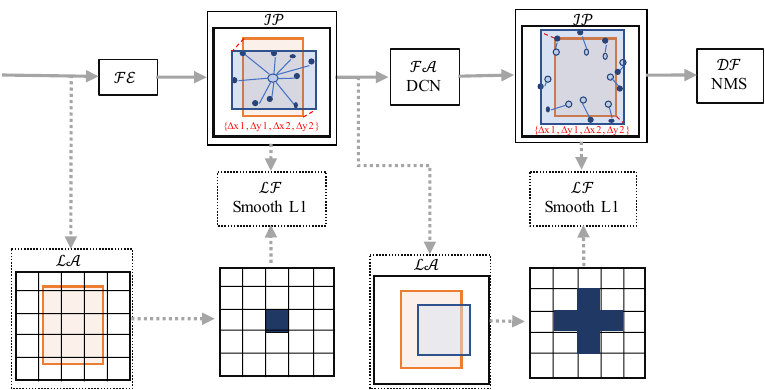}
\caption{Pipelines of a 9-point representation method~\cite{reppoints}. The components are dissected in Sec.~\ref{dissections} to reveal critical aspects for slender object detection.
Dotted boxes and arrows indicate the components only used for training.}
\vspace{-0.1in}
\label{fig:components}
\end{figure}

\begin{table}[tp]
\centering
\caption{Experiments on components making anchor-free detector (FCOS) from an anchor-based detector (RetinaNet).
$\A$: label assign ($\mathcal{LA}$); $\B$: regression space ($\mathcal{IP}$); $\C$: loss function ($\mathcal{LF}$); $\D$: centerness re-weighting~\cite{fcos} ($\mathcal{DF}$ and $\mathcal{LA}$). Details are explained in the first part of Sec.~\ref{anchor-to-anchor-free}.}
\vspace{-0.1in}
\begin{tabularx}{1.0\linewidth}{@{}l*{7}X@{}}
\toprule
\multirow{2}{*}{baseline} & \multirow{2}{*}{w/} & \multicolumn{4}{c}{COCO$^+$ mAR} & COCO   \\ \cline{3-6}
& & all & XS & S & R & mAP \\ \midrule
RetinaNet     &  -          & 48.4    & 23.4 & 37.8      &  53.9   &  36.4     \\
FCOS          &  $\A$-$\D$  & 48.7    & 23.2 & 37.9      &  54.4   &  37.6     \\ \midrule
RetinaNet     &  $\A$       &  37.2   & 15.0 &   26.6    &  42.3   &  30.4     \\
RetinaNet     &  $\A$-$\B$  &  43.6   & 18.6 &   33.1    &  49.3   &  32.2     \\
RetinaNet     &  $\A$-$\C$  &  46.2   & 21.0 &   35.8    &  51.6   &  33.7     \\
RetinaNet     &  $\A$-$\D$  &  48.8   & 22.4 &   37.9    &  54.2   &  37.4     \\
\bottomrule
\end{tabularx}
\vspace{-0.3in}
\label{tab:anchor-dissection}
\end{table}

\NewEnd
\vspace{0.1in}
\noindent\textbf{The controvertible role of anchors}\label{anchor-to-anchor-free}
\vspace{0.1in}

Anchors are once regarded central to many detection systems~\cite{SSD, YOLO} and enable detectors to detect multiple objects at the same location.
However, anchor-based relies on IoU matching between anchors and ground truth boxes for $\mathcal{LA}$, which is shown sub-optimal for slender objects.
This drawback may be partly alleviated by specially designed anchors~\cite{ma2018arbitrary} (see Appendix C for experiments regarding rotated anchors) or bypassed by anchor-free detectors.
Anchor-free detectors~\cite{densebox, cornernet} are alternative approaches that  directly regresses boxes from pixel locations instead of anchors.
This family of detection methods achieve notable success but their properties remains unclear compared with that of anchor-based.

We choose a competitive anchor-free detector FCOS~\cite{fcos} and a classical anchor-based detector RetinaNet~\cite{retina} to reveal the difference between the bifurcated paradigms in essence. We refer readers not familiar with these two methods to Appendix D for an illustration of their architecture.

\New
We show the evaluation of RetinaNet and FCOS in Tab.~\ref{tab:backbones} and Tab.~\ref{tab:anchor-dissection}.
Although FCOS shows advantages in the detection of regular objects and thus keeps ahead in overall performance, the advantage is not consistent for slender objects.
Scrutinizing the evolution from an anchor-based detector to anchor-free detectors like FCOS, we design multiple intermediate variants of which each keeps only one variable to uncover the substantial aspects.
\NewEnd
Excluding implementation details unified by our framework, differences to be addressed exist in $\LF$, $\LA$, and localization paradigm in $\LC$. 
More formally, an anchor-free detector ($\A$) assigns locations inside object boxes as positive, ($\B$)
regresses from pixels rather than from anchors, ($\C$) uses IoU loss as localization target, and specially FCOS ($\D$) adopts centerness score to re-weight loss and confidence at different positions.
Discretely implementing these variants, we show the performance change in Tab.~\ref{tab:anchor-dissection}.


\begin{table}[!tp]
\vspace{-0.2in}
\centering
\caption{Experiments on different box representations. $\E$: nearest label assign ($\mathcal{LA}$), $\F$: 2-point box representation ($\mathcal{DF}$), $\G$: 9-point box representation\cite{reppoints} ($\mathcal{DF}$), $\HH$: feature adaptation ($\mathcal{FA}$). Details are explained in later half of Sec.~\ref{sec:box-representation}.}
\vspace{-0.1in}
\begin{tabularx}{1.0\linewidth}{@{}l*{7}X@{}}
\toprule
\multirow{2}{*}{baseline} & \multirow{2}{*}{w/} & \multicolumn{4}{c}{COCO$^+$ mAR} & COCO   \\ \cline{3-6}
& & all & XS & S & R & mAP \\ \midrule
RetinaNet     &  -  & 48.4 & 23.4 & 37.8    &  53.9   &  36.4     \\
RepPoints     &  $\E$-$\HH$  &  47.0   & 26.2 &   38.1    &  51.0   &  38.4     \\ \midrule
RetinaNet     &  $\E$      &  46.6   & 20.7 &   35.7    &  51.6   &  33.5     \\
RetinaNet     &  $\E$-$\F$  &  42.5   & 22.5 &   33.2    &  46.9   &  32.1     \\
RetinaNet     &  $\E$-$\G$ &  42.2   & 19.6 &   33.0    &  46.6   &  32.1     \\
RetinaNet     &  $\E$-$\HH$  &  46.5   & 25.3 &   37.4    &  50.7   &  38.0     \\
   \bottomrule
\end{tabularx}
\vspace{-0.3in}
\label{tab:box-dissection}
\end{table}

\begin{table*}[!htp]
\vspace{-0.2in}
\centering
\captionof{table}{Validation experiments of self-adaptation. $\checkmark$ and $\times$ indicate models with and without self-adaptation, respectively. The improvements are consistent on stronger models. COCO mAP is evaluated on the validate set, which is usually slightly lower than the dev-test set.}
\vspace{-0.1in}
\begin{tabularx}{1.0\linewidth}{p{1.2cm}p{1.6cm}>{\centering\arraybackslash}p{2.4cm}@{}l*{7}X@{}}
\toprule
  \multirow{2}{*}{Baseline} & \multirow{2}{*}{Backbone}   & \multirow{2}{*}{self-adaptation} &  \multicolumn{4}{c}{COCO$^+$ mAR} & \multirow{2}{*}{COCO mAP} & \multirow{2}{*}{COCO$^+$ mAP} \\ \cline{4-7}
                            &                             &                 & all  &  XS           &   S   &    R     &                           &      \\ \midrule
  \multirow{6}{*}{RetinaNet}&  \multirow{2}{*}{ResNet50}  & $\times$        & 48.4 & 23.4          & 37.8  & 53.9     & 36.4                      & 18.1 \\ \cline{3-9}
                            &                             & $\checkmark$    & 50.1 & 24.4 \GR{1.0} & 39.4  & 55.1     & 39.4 \GR{3.0}             & 19.8 \\ \cline{3-9}
                            &  \multirow{2}{*}{ResNet101} & $\times$        & 50.5 & 23.6          & 39.7  & 55.9     & 38.7                      & 18.7 \\ \cline{3-9}
                            &                             & $\checkmark$    & 52.8 & 27.0 \GR{3.4} & 41.8  & 58.2     & 40.9 \GR{2.2}             & 19.6 \\ \cline{3-9}
                            &  \multirow{2}{*}{ResNeXt101}& $\times$        & 51.1 & 26.7          & 40.5  & 56.2     & 41.9                      & 20.3 \\ \cline{3-9}
                            &                             & $\checkmark$    & 53.9 & 28.3 \GR{1.6} & 43.2  & 58.9     & 43.1 \GR{1.2}             & 21.7 \\ \midrule
  \multirow{6}{*}{FCOS}     &  \multirow{2}{*}{ResNet50}  & $\times$        & 48.6 &    22.2       & 38.6  & 54.5     & 37.6                      & 18.3 \\ \cline{3-9}
                            &                             & $\checkmark$    & 49.2 & 24.5 \GR{2.4} & 38.6  & 54.7     & 39.1 \GR{1.5}             & 19.6 \\ \cline{3-9}
                            &  \multirow{2}{*}{ResNet101} & $\times$        & 50.1 &    24.9       & 39.7  & 55.7     & 40.1                      & 18.9 \\ \cline{3-9}
                            &                             & $\checkmark$    & 50.9 & 26.6 \GR{1.6} & 40.6  & 56.5     & 41.0 \GR{0.9}             & 19.7 \\ \cline{3-9}
                            &  \multirow{2}{*}{ResNeXt101}& $\times$        & 51.9 &    26.5       & 41.4  & 57.8     & 42.5                      & 20.2 \\ \cline{3-9}
                            &                             & $\checkmark$    & 52.6 & 27.5 \GR{1.0} & 42.6  & 57.8     & 43.5 \GR{1.0}             & 20.7 \\ \midrule
  \multirow{6}{*}{RepPoints}&  \multirow{2}{*}{ResNet50}  & $\times$        & 47.0 & 26.2          & 38.1  & 51.0     & 38.4                      & 18.4 \\ \cline{3-9}
                            &                             & $\checkmark$    & 50.1 & 28.4 \GR{2.2} & 40.4  & 54.4     & 39.4 \GR{1.3}             & 19.9 \\ \cline{3-9}
                            &  \multirow{2}{*}{ResNet101} & $\times$        & 49.8 & 27.1          & 39.5  & 54.9     & 39.7                      & 19.8 \\ \cline{3-9}
                            &                             & $\checkmark$    & 50.1 & 28.7 \GR{1.6} & 40.7  & 55.0     & 40.6 \GR{0.9}             & 20.3 \\ \cline{3-9}
                            &  \multirow{2}{*}{ResNeXt101}& $\times$        & 51.5 & 27.6          & 41.4  & 55.8     & 42.2                      & 21.1 \\ \cline{3-9}
                            &                             & $\checkmark$    & 52.1 & 30.0 \GR{2.4} & 42.5  & 56.8     & 42.7 \GR{0.5}             & 21.5 \\
   \bottomrule
\end{tabularx}
\vspace{-0.2in}
\label{tab:sa-validation}
\end{table*}

\New
We can draw clear conclusion from the experiments.
The removal of anchor-based assignment brings such significant performance gap between the two paradigms that all other components are introduced to fulfill the rift.
We also notice the impact of other attempts remain fairly trivial until $\mathcal{LA}$ is improved. It leads to the conclusion that \textbf{the essential role of anchors is in label assignment}.

In comparison with the difference in $\LA$ that bifurcates the anchor-free detector from anchor-based RetinaNet with $6\%$ mAP, the effect of anchors for regression is minor. Other interesting phenomenon includes that the improvement of IoU loss is more significant than expected~\cite{giou}, especially for slender objects. We hypothesis the cause is two fold: (1) The anchor-free $\mathcal{LA}$ guarantees the predicted bounding box overlaps with the ground truth, while anchor-based doesn't. (2) Inaccurate predictions around box borders are less punished by IoU loss and thus are less likely to affect predictions at center area during NMS. These properties are enhanced by FCOS using centerness to achieve finally better performance than anchor-based baseline. It is also possible to utilize this property in slender object detection as a trade-off strategy. The details are provided in Appendix D. \NewEnd

\vspace{0.1in}
\noindent\textbf{The impact of box representation}~\label{sec:box-representation}
\vspace{0.1in}

\New
In addition to anchors, we suppose box representations~\cite{reppoints, centernet} that are closely related to object slenderness would help this problem.
Works in this direction modify the regression targets and form bounding boxes from other representations during $\mathcal{DF}$.
\NewEnd
Among these methods, RepPoints~\cite{reppoints} stands out due to its promising improvements and novel box representation.
We re-implement it inside our analytical framework to conduct fair experiments for mining valuable components in terms of slender objects.

The model architecture of RepPoints is illustrated in Fig.~\ref{fig:components}. 
It turns box representation into a set of points, specifically 9 in their experiments, which forms a pseudo box using simple processing such as min-max. 
Furthermore, the 9-point representation coordinates with Deformable Convolution (DCN) by forming its offsets. Following the proposed $\FA$ layer, an extra $\LC$ stage is performed to refine the initial localization results.
Applying our framework, RepPints is distinguishable in several aspects: ($\E$) label assignment by assigning the nearest location to box centers as positive; ($\F$) 2-point representation instead of anchor-based representation; ($\G$) the proposed 9-point representation and pseudo box formation; and ($\HH$) supervised feature adaption integrating localization results with DCNs. 

\New
Despite the remarkable performance validated by our experiments, the potential root is surprising: In comparison with 2-point representation, the 9-point representation demonstrates no advantages with the same $\mathcal{LA}$.
Note the 2-point representation is equivalent to the regression strategy of anchor-based methods like RetinaNet.
We claim that different \textbf{box representations are comparable in object localization} and it is further verified in detailed experiments in the next section.
In contrast, the supervised \textbf{feature adaptation is critical for slender object detection} and substantially improves mAR on XS objects.

Through experiments in Tab.~\ref{tab:box-dissection}, we recognize $\mathcal{FA}$ as the potential root of advantages in slender object detection.
However, the supervision that forces DCN to sample features around the borders (see Fig.~\ref{fig:components}) contradicts the property of slender objects that locations near border are more likely to be background.
This thinking inspires us to propose a novel feature adaptation strategy in next section.
\NewEnd

\section{Improving Slender Object Detection} \label{improvements}

\New
In Sec~\ref{component-inspectino}, we dispute the conjecture that assumes anchors or box representations are central and recognize that feature adaptation lead to improvements of slender object detection.
On the other hand, inspections also reveal ambiguities in understanding the effects.
The supervised feature adaptation accordingly constrains the offsets of DCN by exploiting an intermediate localization stage.
Since the refinement of feature and the basis of the final prediction are coupled in sample points, the necessity is arguable.
In this section, we first propose a self-adaptation strategy of features that significantly improves the detection accuracy of slender objects and then demonstrate its properties with extensive ablation experiments using our analytical framework.
\NewEnd

\subsection{Self-Adaptation of Features}\label{supervised-feature-adaption}

\begin{table}[tp]
\vspace{-0.1in}
\centering
\caption{Ablation study on feature adaption. Modifications $\I$ to $\LL$ are introduced in Sec.~\ref{sec:ablation-for-sa}, and the original and our improved versions of the mentioned methods are marked with $*$ and $\dag$, respectively. The mAP and mAR (XS) are evaluated on COCO and COCO$^+$, where XS stands for extra slender objects.}
\vspace{-0.1in}
\begin{tabularx}{1.0\linewidth}{@{}l*{5}X@{}}
\toprule
\# & baseline & modules & mAP  & mAR(XS)\\ \midrule
M1      & RepPoints & $\I$+$\LL$   & 37.9 & 24.9\\
M2$^*$  & RepPoints & $\I$+$\J$+$\LL$ & 38.4 & 25.1\\
M3      & RepPoints & $\I$+$\J$   & 38.1 & 27.6\\
M4$^\dag$     & RepPoints & $\I$+$\K$   & \textbf{39.4} & \textbf{28.4}\\ \midrule
M5$^*$ & FCOS      & -     & 37.6 & 23.2\\
M6      & FCOS      & $\I$     & 37.7 & 23.9\\
M7      & FCOS      & $\I$+$\J$   & 39.1 & 24.6\\
M8$^\dag$ & FCOS      & $\I$+$\K$   & 39.0 & 24.5\\ \midrule
M9$^*$ & RetinaNet & -     & 37.4 & 23.4\\
M10     & RetinaNet & $\I$    & 38.4 & 24.2\\
M11     & RetinaNet & $\I$+$\J$   & 39.1 & 24.3\\
M12$^\dag$    & RetinaNet & $\I$+$\K$   & \textbf{39.4} & 24.4\\ \bottomrule
\end{tabularx}
\vspace{-0.1in}
\label{fig:improvement}
\vspace{-0.1in}
\end{table}

\New
Inspired by analyses in Sec.~\ref{sec:box-representation}, we focus on $\mathcal{FA}$ to improve slender object detection. 
The concept of feature adaptation is kept and we generalize it to fit into the nature of slender objects. Instead of manually supervise the feature adaptation to sample features from borders, we deploy feature adaptation without explicitly constraining the sampling points, namely self-adaptation. Self-adaptation of features is characterized by two aspects. (1) It uses a DCN layer whose produced feature is supervised by an initial initial box regression task similar to \cite{guided-anchor} and \cite{reppoints}. Consequently, the feature is refined for classification. 
(2) The adaptation is not directly constrained by annotations to be optimal for slender objects, in contrast to \cite{reppoints}.
We use regular 2-point representation of object boxes for the initial box regression, since our experiments show its descriptive capacity is comparably powerful.

We conduct extensive experiments on different baselines with different backbones to validate the improvement of self-adaptation, especially for slender objects. The results are shown in Tab.~\ref{tab:sa-validation}. Self-adaptation that can be used as an enhancement in different detection paradigms brings significant and consistent improvement to detection of slender objects.
Focusing on extra slender objects (XS in the table), the evaluation of self-adaptation demonstrates improvements over all baselines, despite their original properties in slender object detection.
The improvement is even more remarkable for stronger backbones, indicating that self-adaptation is suitable for features with higher quality.
Although COCO is biased against slender objects (shown in Sec.~\ref{sec:assessment}), the improvements in slender objects also reflect to the overall mAP on COCO.
\NewEnd

\begin{figure}[!tbp]
\vspace{-0.1in}
\centering
\begin{subfigure}[b]{0.32\columnwidth}
    \centering
    \includegraphics[width=\textwidth]{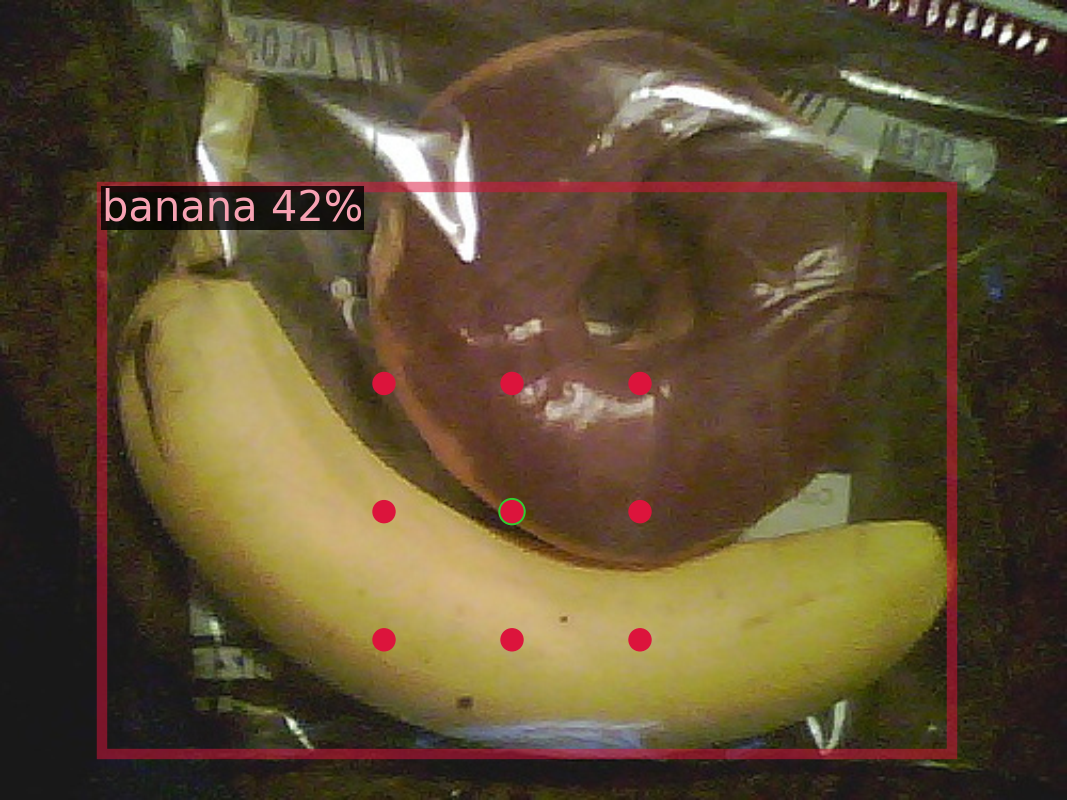}
    \caption{convolution}
\end{subfigure}
\begin{subfigure}[b]{0.32\columnwidth}
    \centering
    \includegraphics[width=\textwidth]{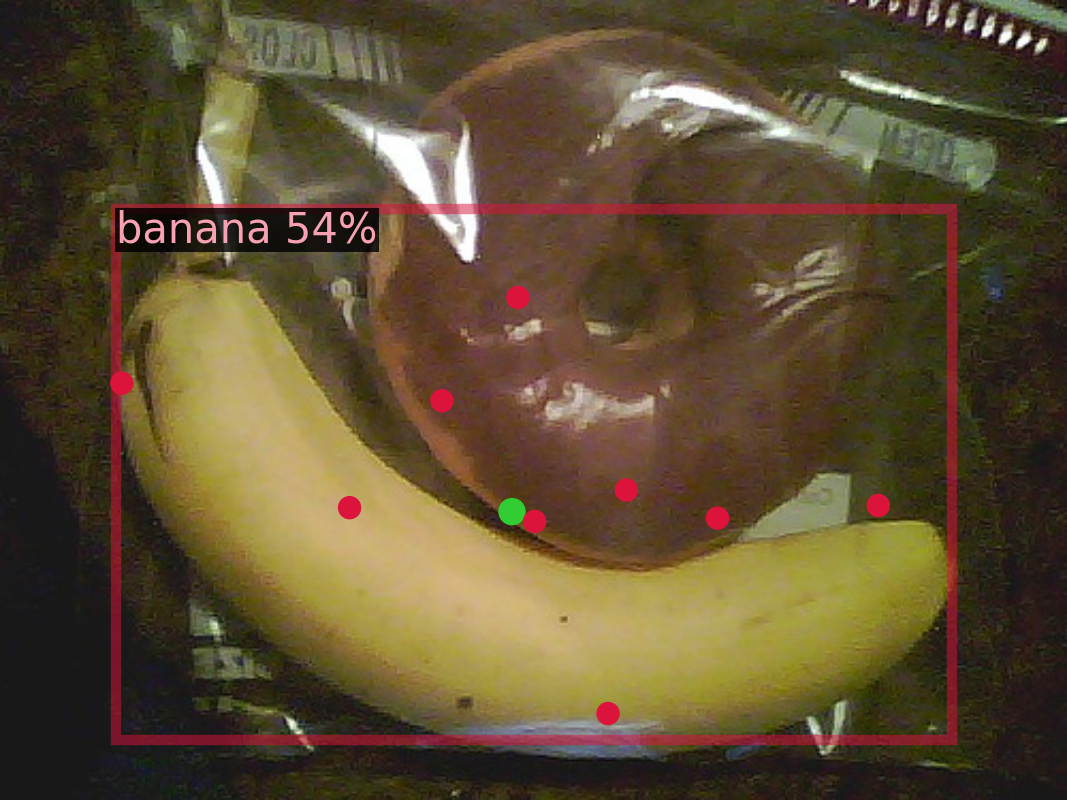}
    \caption{reppoints\cite{reppoints}}
\end{subfigure}
\begin{subfigure}[b]{0.32\columnwidth}
    \centering
    \includegraphics[width=\textwidth]{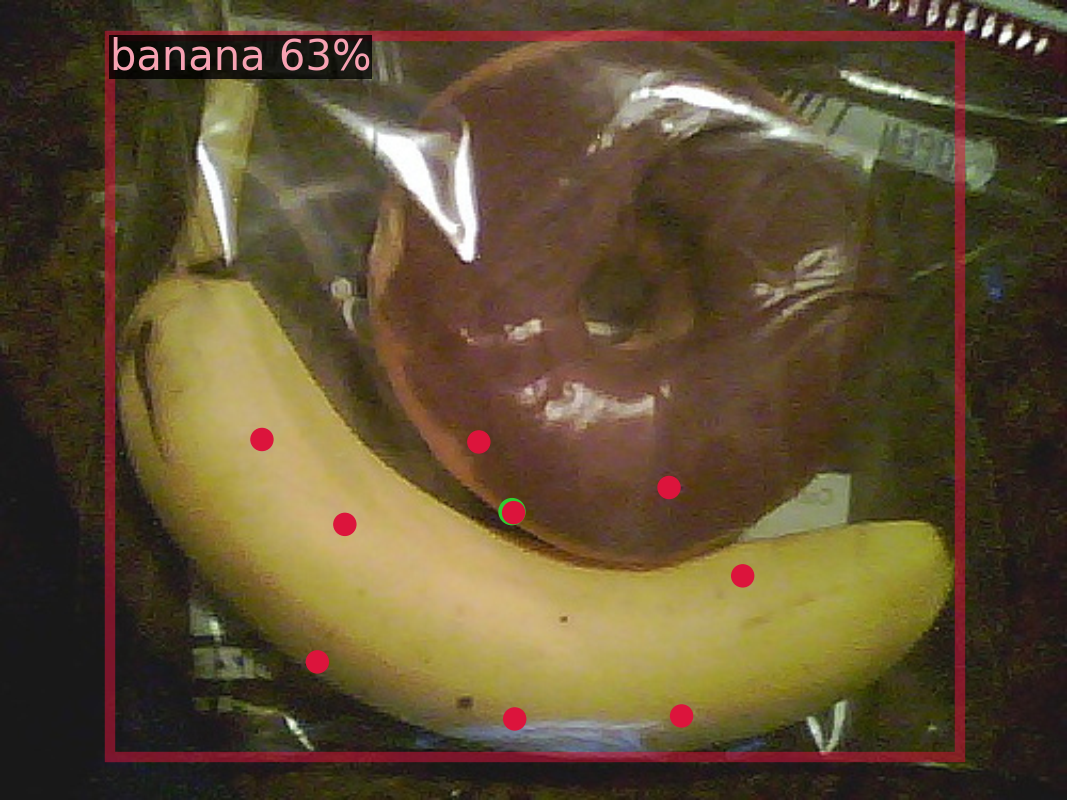}
    \caption{self-adaption}
\end{subfigure}
\vspace{-0.1in}
\caption{Visualization of different feature adaption strategies. The sampling points are marked as red points relative to the green points (See more details in Sec.~\ref{supervised-feature-adaption}).}
\vspace{-0.1in}
\label{fig:visualization}
\end{figure}

\subsection{Ablation Study} \label{sec:ablation-for-sa}

In this section we use our proposed analytical framework to conduct ablation study in self-adaptation. 
Rigorously controlling variables, we experiment on following variants to show the impact of each of the modifications that makes self-adaptation.
($\I$) an initial localization in addition to the final results presented by RepPoints; ($\J$) constraining offsets of DCN using the initial object localization; ($\K$) offsets adaptively learned from the features; and ($\LL$) a residual-manner final localization that infers upon the initial localization.
Modules ($\I$) and ($\LL$) follow the design of RepPoints~\cite{reppoints}.

\New
The experiments generally verify self-adaptation as the root of the improvements on slender object detection demonstrated in Tab~\ref{tab:sa-validation}.
Several approaches use residual prediction that refines the prediction based on the initial regression results~\cite{cascade, cascadeface, reppoints}. However, our ablation ($\LL$) shows it brings unnecessary drift to slender object detection. Moreover, we are able to safely remove the direct supervision of feature adaptation since the experiments ($\mathbb{K}$) verifies our conjecture that this supervision contradicts the property of slender objects. In Fig.~\ref{fig:visualization}, we visualize the sampling points of self-adaptation and other strategies. By forcing the sampling points to concentrate on the foreground of objects and to avoid the interference of background, self-adaption refines the feature to capture foreground objects and achieves better detection.
\NewEnd

\section{Conclusion}

In this paper, we investigate an important yet long-overlooked problem of slender object detection. A comprehensive framework is established  for dissecting and comparing different object detection methods as well as their components and variants. Based on this framework, a series of key observations and insights is obtained. Furthermore, we have proposed an effective strategy for significantly improving the performance of slender object detection.

\appendix
\section{Modifications Lookup Table}

\begin{table*}[tp]
\centering
\caption{Lookup table for our model assessment.}
\vspace{-0.1in}
\begin{tabularx}{1.0\linewidth}{@{}l*{3}X@{}}
\toprule
$\#$         &  Stage           & Description             & Reference\\ \midrule
$\A$         &  $\mathcal{LA}$  & Assign in-object locations as positive  & FCOS~\cite{fcos} \\
$\B$         &  $\mathcal{IP}$  & Regression from object center             & FCOS~\cite{fcos} \\
$\C$         &  $\mathcal{LF}$  & IoU Los     & IoUNet~\cite{jiang2018acquisition}, FCOS~\cite{fcos} \\
$\D$         &  $\mathcal{LA}, \mathcal{DF}$  & Center-prior          & FCOS~\cite{fcos} \\ \midrule
$\E$         &  $\mathcal{LA}$  & Assign object center as positive          & RepPoints~\cite{reppoints} \\
$\F$         &  $\mathcal{IP}$  & 2-point box representation          & RetinaNet~\cite{retina}, RepPoints~\cite{reppoints}\\
$\G$         &  $\mathcal{IP}$  & 9-point box representation          & RepPoints~\cite{reppoints} \\
$\HH$        &  $\mathcal{FA}$  & Supervised feature adaptation          & RepPoints~\cite{reppoints} \\ \midrule
$\I$         &  $\mathcal{FA}$  & Initial localization prediction          & RepPoints~\cite{reppoints} \\
$\J$         &  $\mathcal{FA}$  & Supervised DCN          & RepPoints~\cite{reppoints} \\
$\K$         &  $\mathcal{FA}$  & Self-adaptation of features          & - \\
$\LL$        &  $\mathcal{IP}$  & Residual localization prediction          & RepPoints~\cite{reppoints} \\
\bottomrule
\end{tabularx}
\vspace{-0.2in}
\label{tab:lut}
\end{table*}

To help understand the various experiment inconsistency we address in this paper, we prove a lookup table for the ablation modifications involved in the paper in Tab.~\ref{tab:lut}. We also attached our code in supplementary materials.

\section{Pseudo Mask Generation}

We use images containing slender objects from Objects365~\cite{obj365} Dataset to complement COCO.
As there are 365 different categories in Objects365, we map objects categories that appear in COCO into the COCO$^+$ dataset to be mixed with the original COCO.

\begin{figure}[tbp]
    \centering
    \includegraphics[width=0.9\columnwidth]{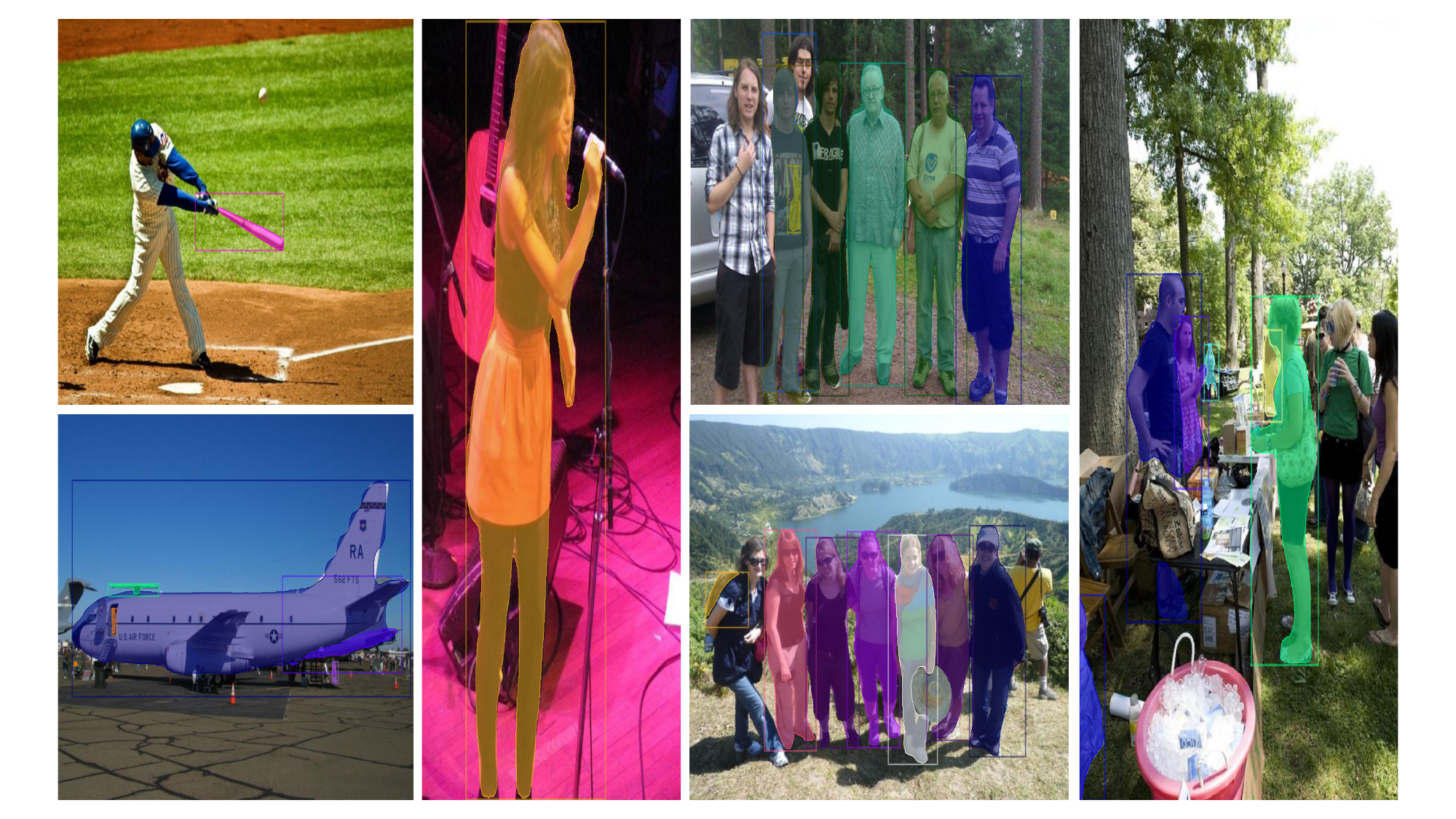}
    \caption{Pseudo masks generated by Cascaded RCNN~\cite{cascade} for $COCO^+$. The boxes are the ground-truth labels and the masks are accordingly generated. Only masks for slender objects are visualized.}
    \label{fig:pseudo_masks}
\end{figure}

For slenderness estimation, we use a top-performance instance segmentation model from Detectron2~\cite{wu2019detectron2}, an implementation of \cite{cascade} with a ResNeXt152 backbone.
Benefiting from the pre-trained models of Dectectron2, we do not need to re-train the segmentation model.
We use no test time augmentation during inference, and the resolution of the input images during testing is fixed to (800, 1333).
As mentioned in Section 2.2 in our paper, the ground truth bounding boxes are used as proposals while generating the masks.
Consequently, the produced pseudo masks are accurate enough and reliable to measure the slenderness of objects.
Some examples of the generated pseudo masks of slender objects are shown in Figure.~\ref{fig:pseudo_masks}.

\section{Rotated Bounding Box Evaluation}

\begin{figure}[tbp]
\begin{subfigure}[b]{0.9\columnwidth}
    \centering
    \includegraphics[width=\textwidth]{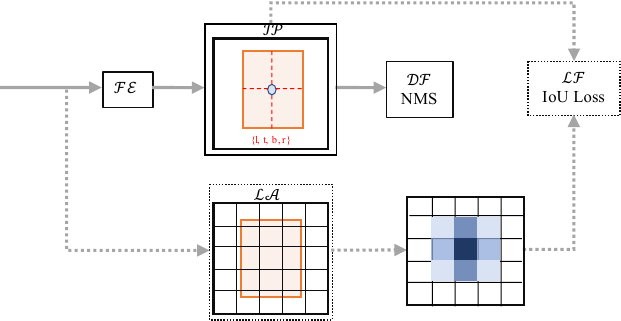}
    \caption{FCOS}
\end{subfigure}
\begin{subfigure}[b]{0.9\columnwidth}
    \centering
    \includegraphics[width=\textwidth]{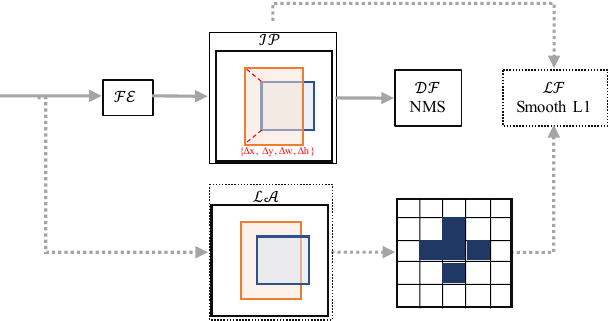}
    \caption{RetinaNet}
\end{subfigure}
\vspace{-0.1in}
\caption{Pipelines of methods chosen for problem dissection.
Dotted boxes and arrows indicate the components only used for training.}
\vspace{-0.2in}
\label{fig:components-anchor}
\end{figure}

In scenarios where the rotation is required for meaningful detection, \eg scene text detection in the wild and ship detection in aerial images, rotated boxes are applied to represent objects. As another challenging problem to be solved, the detection of oriented objects is faced with particular difficulties due to discontinuity in angles~\cite{yang2019scrdet}. Research towards this problem is making notable progress~\cite{yang2020arbitrary, zhou2020objects}, while challenges persist. We provide baseline evaluation of slender objects on both COCO and COCO+, to open up new opportunities in this field.

Similar to slenderness estimation from COCO, we approximate the angle of rotated bounding box using $90$-degree box representation shown in \cite{yang2020arbitrary}. Inspired by \cite{ma2018arbitrary}, we extend FasterRCNN to support rotated bounding boxes. The evaluation results is shown in Tab.~\ref{tab:rotated}. The shown mAP and mAR are evaluated on COCO and COCO$^+$ validation set, separately.

\section{Architecture of RetinaNet and FCOS}

In Sec.\ 3 of the paper, we use FCOS and RetinaNet as baseline models to conduct our analysis regarding the role of anchors. In case readers are not familiar with these two methods, we illustrate their model architecture in Fig.~\ref{fig:components-anchor}. RetinaNet~\cite{retina} is a classical and effective detector that produces localization and classification results per location. Both of its $\mathcal{IP}$ and $\mathcal{LA}$ rely on manually designed anchors. The $\mathcal{IP}$ is targeted at predicting $\Delta$ from an anchor to its matched maximum IoU object. As of $\mathcal{LA}$, the IoU between anchor and object bounding boxes is used to assign labels for each anchor at each location. 
FCOS~\cite{fcos} performs object detection in a different manner, namely anchor-free detector. It assigns locations inside object bounding boxes with corresponding object label, and the regression target is to predict distances from a location to box boundaries. As we demonstrated in the paper, the centerness score that underlines the center areas of objects, is key to the effectiveness of FCOS. The localization loss and prediction scores are re-weighted by the centerness scores.

\begin{table}[tp]
\centering
\caption{Rotated box evaluation on validation datasets. IoU is now mesured between rotated bounding boxes to match prediction and ground truth.} \label{tab:rotated}
\vspace{-0.1in}
\begin{tabularx}{1.0\linewidth}{@{}l*{6}X@{}}
\toprule
\multirow{2}{*}{backbone} & \multicolumn{4}{c}{COCO$^+$ RBox mAR} & RBox   \\ \cline{2-5}
&  all & XS & S & R & mAP  \\ \midrule
ResNet-50     &  37.5   &   18.3    &   27.6  &  41.5 &  27.7     \\
ResNet-101    &  38.9   &   19.4    &  29.6   &  42.6 &  29.8     \\
   \bottomrule
\end{tabularx}

\end{table}

\section{Slenderness Prior}

\begin{table}[tp]

\centering
\caption{Validation of the slenderness prior. $\FA$ is module $\K$ presented in the paper, and SP is short for slenderness prior.}
\vspace{-0.1in}
\begin{tabularx}{1.0\linewidth}{@{}l*{8}X@{}}
\toprule
\multirow{2}{*}{baseline} & \multirow{2}{*}{$\FA$} & \multirow{2}{*}{SP} & \multicolumn{4}{c}{COCO$^+$ mAR} & COCO   \\ \cline{4-7}
& & & all & XS & S & R & mAP\\ \midrule
FCOS   &  &   &  48.9   & 23.9
&   38.6    &  54.5   &  37.7     \\
FCOS   & \checkmark &  & 49.4
&   24.2    &  39.0   &  55.1 &  39.0     \\
FCOS   & \checkmark &  \checkmark  &  49.7   & 26.3
&   40.0    &  54.2   &  38.4     \\
   \bottomrule
\end{tabularx}

\label{tab:slender-prior}
\end{table}

We have mentioned in the paper that the center prior is recognized as a central component for anchor-free detectors and its extension makes a trade-off strategy between regular and slender objects.
Center prior suppresses spurious prediction that is distant from the object centers by re-weighting using centerness scores defined by 
\begin{equation}
    centerness = (\frac{\min(l, r)}{\max(l, r)} \times \frac{\min(t, b)}{\max(t, b)})^{\frac{1}{2}}.
\end{equation}
$l, r, t, b$ are the distance to the left, right, top, and bottom border of the bounding box, respectively.
With the geometric mean, the decay is slower on the long sides of slender objects but faster on the short sides, causing insufficient training of locating slender objects. Naturally, we extend the formula to
\begin{equation}
    centerness^* = (\frac{\min(l, r)}{\max(l, r)} \times \frac{\min(t, b)}{\max(t, b)})^{s},
\end{equation}
where $s$ is the slenderness of objects.
It favors slender objects that are challenging for precise detection and fasten the score decay of regular objects.

To validate the effectiveness of slenderness prior, we perform experiments using the baseline model of FCOS (M5) and its variant with self-adaption (M8) introduced in the paper.
As the results in Tab.~\ref{tab:slender-prior} demonstrate, this natural extension significantly improves the detection mAR for slender objects, with an acceptable sacrifice of the mAP for R objects.
Despite an mAR degradation for R objects, the mAR of XS and S improve 2.1\% and 1.0\%, respectively.
It indicates that the slenderness prior is a favorable trade-off between slender and regular objects, as the overall mAR reaches 49.7\%.

{\small
\bibliographystyle{ieee_fullname}
\bibliography{egbib}
}
\end{document}